\documentclass[sigconf,nonacm]{acmart}
\usepackage{enumitem}

\usepackage{balance} 

\title{LLM-based Multi-Agent Systems: \\Techniques and Business Perspectives}

\author{Yingxuan Yang}
\email{zoeyyx@sjtu.edu.cn}
\affiliation{%
  \institution{Shanghai Jiao Tong University}
 \city{Shanghai}
 \country{China}
}

\author{Qiuying Peng}
\email{qypeng.ustc@gmail.com}
\affiliation{%
  \institution{OPPO Research Institute}
 \city{Shenzhen}
 \country{China}
}

\author{Jun Wang}
\email{junwang.lu@gmail.com}
\affiliation{%
  \institution{OPPO Research Institute}
 \city{Shenzhen}
 \country{China}
}

\author{Ying Wen}
\email{ying.wen@sjtu.edu.cn}
\affiliation{%
  \institution{Shanghai Jiao Tong University \& SII}
  \city{Shanghai}
  \country{China}
}

\author{Weinan Zhang}
\email{wnzhang@sjtu.edu.cn}
\affiliation{%
  \institution{Shanghai Jiao Tong University \& SII}
 \city{Shanghai}
 \country{China}
}
\authornote{Corresponding Author}

\begin{abstract}
In the era of (multi-modal) large language models, most operational processes can be reformulated and reproduced using LLM agents. The LLM agents can perceive, control, and get feedback from the environment so as to accomplish the given tasks in an autonomous manner. Besides the environment-interaction property, the LLM agents can call various external tools to ease the task completion process. The tools can be regarded as a predefined operational process with private or real-time knowledge that does not exist in the parameters of LLMs. As a natural trend of development, the tools for calling are becoming autonomous agents, thus the full intelligent system turns out to be a LLM-based Multi-Agent System (LaMAS). Compared to the previous single-LLM-agent system, LaMAS has the advantages of i) dynamic task decomposition and organic specialization, ii) higher flexibility for system changing, iii) proprietary data preserving for each participating entity, and iv) feasibility of monetization for each entity. This paper discusses the technical and business landscapes of LaMAS. To support the ecosystem of LaMAS, we provide a preliminary version of such LaMAS protocol considering technical requirements, data privacy, and business incentives. As such, LaMAS would be a practical solution to achieve artificial collective intelligence in the near future.
\end{abstract}

\keywords{LLM-based Multi-Agent System, large language model, data privacy, monetization}

\newcommand{\BibTeX}{\rm B\kern-.05em{\sc i\kern-.025em b}\kern-.08em\TeX}

\newcommand{\minisection}[1]{\noindent\textbf{#1.}}

\begin{document}


\pagestyle{fancy}
\fancyhead{}


\maketitle 






\section{Background and Trend}
The development of Large Language Models (LLMs) \cite{openai2024gpt4technicalreport} marks a key advancement in artificial intelligence.  These models have transformed from simple text processors to sophisticated systems capable of reasoning, understanding multimodal inputs, and making autonomous decisions \cite{wang2024openr}. Such developments have enabled the emergence of AI agents powered by LLMs\footnote{For presentation brevity, in this paper, the multi-modal LLM concept \cite{caffagni2024r} is merged into the LLM concept.}, which can adapt to diverse tasks, comprehend context, and interact with their environments autonomously \cite{agenticinformationretrieval, trad}.

A critical transition in LLM capabilities is their evolution from passive tools that merely respond to commands to active agents capable of independent decision-making and action-taking \cite{schick2023toolformerlanguagemodelsteach, hammer}. Initially, LLMs were primarily used for single-purpose tasks, such as text generation or analysis. However, recent advances have equipped them to interact with graphical user interfaces (GUIs) and perform complex operations such as web browsing, app navigation, and system control \cite{zhang2024largelanguagemodelbrainedgui}. Beyond these capabilities, modern LLMs have transformed into autonomous agents that dynamically select and use tools based on contextual requirements. This evolution highlights their dual nature: they not only utilize tools but can also function as tools within modular systems, enabling the creation of multi-agent architectures where agents collaborate to solve complex problems.

The rise of \textbf{LLM-based Multi-Agent Systems (LaMAS)} marks a significant leap in AI applications \cite{chen2024internetagentsweavingweb,liu2024autonomousagentscollaborativetask}. Although such systems may require greater computational resources compared to single-agent approaches, they offer crucial advantages that justify this trade-off: inherent fault tolerance through agent redundancy, natural task decomposition without explicit workflow design, and organic specialization in complex problem-solving. When one agent fails in a multi-agent system, others can seamlessly continue operations, providing robust reliability that centralized systems cannot match. Moreover, while single-agent systems demand careful orchestration of execution workflows for each task type, multi-agent systems naturally emerge with collaborative specialization patterns, allowing each agent to focus on its core competencies within the larger system architecture.

Recognizing these benefits, researchers have developed LaMAS frameworks to enable complex task collaboration \cite{guo2024largelanguagemodelbased, chen2024internetagentsweavingweb,liu2024autonomousagentscollaborativetask, fourney2024magenticonegeneralistmultiagentsolving, ghafarollahi2024sciagentsautomatingscientificdiscovery,chen2023agentverse,chatdev,hong2024metagpt,wu2023autogen,xie2023openagents, li2023metaagentssimulatinginteractionshuman,gao2024agentscopeflexiblerobustmultiagent,zhugegptswarm,trirat2024automlagentmultiagentllmframework,fu2024msiagentincorporatingmultiscaleinsight,li2024agentorientedplanningmultiagentsystems,chan2023chatevalbetterllmbasedevaluators}. Beyond traditional paradigms like SaaS, PaaS, and IaaS, LaMAS introduces a novel approach by seamlessly integrating intelligent agents into cloud ecosystems. This framework supports the deployment of specialized agents capable of collaboration while maintaining data privacy and security. Furthermore, it establishes a marketplace for agent monetization, allowing users to customize and combine agent services according to their needs. The system architecture emphasizes modular design, standardized communication protocols, and robust security measures, fostering sustainable innovation.

\minisection{Incentivization via Monetization Mechanisms}
Just like the Internet applications are highly incentivized to connect to the Internet, the agents in a LaMAS are also highly incentivized based on monetization mechanism design. First, the experience data generated from interacting within a LaMAS is crucial for training well-functional agents. In LaMAS, the agents receive the task instructions from upstream agents, perform inner-agent reasoning and tool usage, send task instructions to downstream agents and acquire their returned information, and obtain the final task accomplishment results. Such experience data is more valuable and of a larger volume than a single agent just connecting to the users. 
Second, similar to Internet monetization via online advertising, there will be a monetization mechanism over the LaMAS. Specifically, for each accomplished task that is assigned with a business value, e.g., the user books a hotel or purchases an item, there will be a promotion fee from the merchant provided to the engaged team of agents in the LaMAS and the credit allocation mechanism can be built by the participation or the essential contribution to the task accomplishment. As such, the entity behind each agent has the essential motivation to build a highly intelligent agent connecting to the LaMAS.

\minisection{Entity's Responsibility based on Agent Intelligence}
For the Cable or 4/5G Internet, each entity behind an Internet service, e.g., the company, institute, or team, is responsible for maintaining a stable function running and connection to the Internet. If its server crashes or the connection is disabled, the other services depending on its service will be highly influenced, thus the entity should take charge of the influence it makes.
Analogously, in the LaMAS ecosystem, the entity behind each agent has the responsibility to make the ecosystem run smoothly and intelligently. First, inherited from the Internet services, each agent needs to support a stable function running and connection to the Internet. Second, more importantly, the intelligence provided by the agent must meet or exceed predefined standards, as the low intelligence offered by an agent would possibly make the whole LaMAS less functional for accomplishing intelligence tasks.

In this paper, we show our perspectives on LaMAS by discussing its technical and business landscapes. We will show the key AI technical aspects, including system architectures, collaboration protocols, agent training methods, and business aspects, including data privacy-preserving and monetization via traffic and intelligence.
With these analyses, it is worth expecting that LaMAS will form a new technical business paradigm in the coming few years.

\section{Key AI Technical Aspects}

\subsection{Architecture of LLM Agents}
The architecture of LLM-based AI agents consists of several interrelated components essential for autonomous operation and intelligent interaction. At its core, this architecture is designed to effectively process inputs, maintain contextual relevance, make informed decisions, and generate appropriate responses.

The interaction wrapper serves as the principal interface through which the agent interacts with its environment and other agents. This component manages the flow of incoming and outgoing communications. It adapts to various input modalities, and standardizes them for internal processing. The interaction wrapper implements protocol-specific adaptations to ensure seamless integration with various communication standards. This approach preserves the internal consistency of the agent’s operations.

Memory management is pivotal to the architecture. It includes both short-term working memory and long-term episodic storage. The short-term memory buffer retains immediate context and recent interactions, facilitating conversational coherence. Meanwhile, the long-term memory system archives significant experiences and learned patterns, enabling the agent to adapt its responses based on historical interactions and enhancing decision-making capabilities in contextually rich scenarios.

The cognitive functionality of the architecture is currently underpinned by Chain-of-Thought (CoT) reasoning \cite{wei2023chainofthoughtpromptingelicitsreasoning, yao2023reactsynergizingreasoningacting}. This structured reasoning framework decomposes complex tasks into manageable logical steps, thereby facilitating clarity and thoroughness in problem-solving. The CoT mechanism enables the agent to articulate intermediate reasoning states, verify logical consistency, and engage in self-correction through a systematic analysis of its reasoning processes.

Additionally, to enhance the agent's operational capacity beyond natural language processing, the tool integration framework is necessary \cite{schick2023toolformerlanguagemodelsteach, patil2023gorillalargelanguagemodel, hammer}. This subsystem discovers and registers tools, maps of parameters between natural language commands and tool APIs, monitors execution, handles errors, and interprets results. It ensures the effective integration of external functionalities into its decision-making processes.

The architecture also features a sophisticated routing mechanism that governs the connections with neighboring agents. This component is instrumental in facilitating dynamic neighbor discovery, making capability-based routing decisions, balancing loads across the agent network, and enforcing policy-based access control. Such networking capabilities are vital for ensuring efficient communication and collaboration within multi-agent systems.

Furthermore, the architecture incorporates feedback loops that enable continuous learning and adaptation. These loops facilitate the processing of interaction outcomes, allowing the agent to update its internal models and refine its decision-making strategies based on experiential learning. The integration of these architectural elements not only establishes a robust foundation for the autonomous operation of LaMAS but also significantly enhances their collaborative capabilities within multi-agent systems.

\begin{figure}[t]
    \centering
    \includegraphics[width=\linewidth]{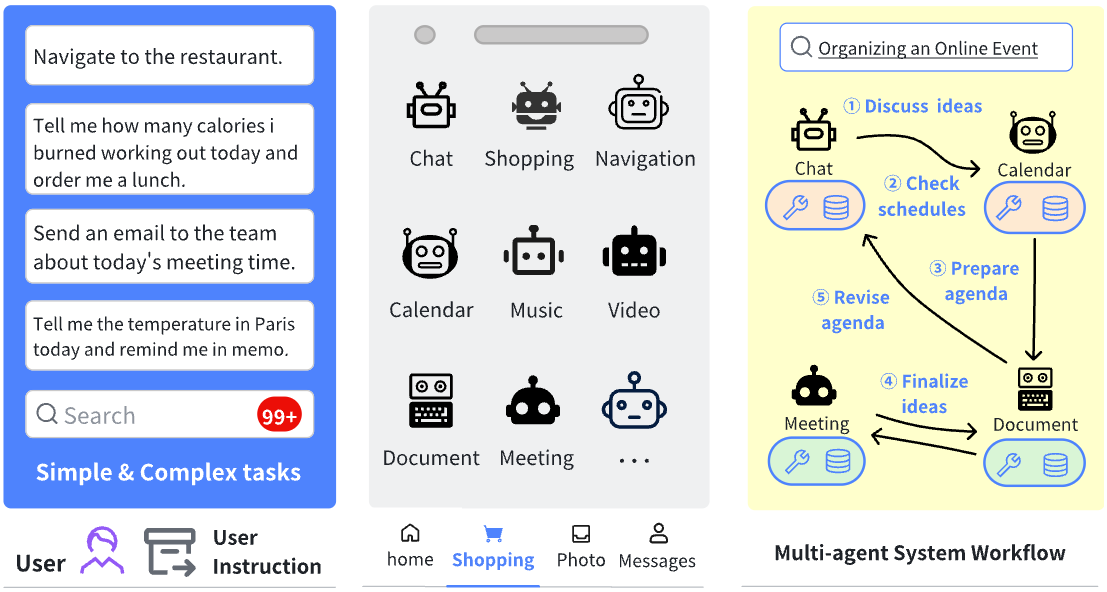}
    \caption{Illustration of LaMAS.}
    \label{fig:mas_illustration}  
    \vspace{-0.5cm}
\end{figure}

\subsection{Mechanisms and Architectures of LaMAS}
As a multi-agent system, the design of mechanisms and architectures of a LaMAS is crucial for its success. Roughly, according to the coordination form of a MAS, there are three major architectures of a LaMAS.

First, for fully centralized architectures, the whole system has full control of the engaged agents, which is a very high requirement, then centralized training with centralized execution methods can be used, and the agents will act with high coordination. In practice, such methods can be applied only when the agents are the applications developed over an OS-liked platform and grant the data and control access to the platform.

Second, for decentralized architectures with global credit allocation, the whole system cannot fully control the engaged agents but can allocate credit to each one for each accomplished task, then centralized training with centralized execution methods can be applied. This is much more practical since each agent (and the entity behind it) does not have to grant the data or control access to the platform. Also, the platform can still incentivize the engaged agents to improve their collaboration performance by allocating credit to the team.

Third, for fully decentralized architecture, i.e., there is no access to data or control for each engaged agent and no credit allocation for the platform, the agents in the system will need to find their own way to collaborate with others and improve themselves. In such a case, the mechanism design will be much important from the beginning.


\subsection{Protocols of Agent Interaction}
The LLM-based Multi-Agent System (LaMAS) framework necessitates sophisticated interaction protocols to facilitate effective agent collaboration. These protocols must bridge the gap between traditional structured formats and natural language understanding, addressing unique challenges posed by LLM-based agents' probabilistic decision-making and emergent capabilities \cite{marro2024scalablecommunicationprotocolnetworks}.

\minisection{Core Challenges and Key Issues}
The development of LaMAS protocols presents several fundamental challenges in protocol effectiveness measurement, behavioral diversity optimization, and non-transitive interaction management. 
\vspace{-0.2cm}
\begin{figure}[h]
    \centering
    \includegraphics[width=\linewidth]{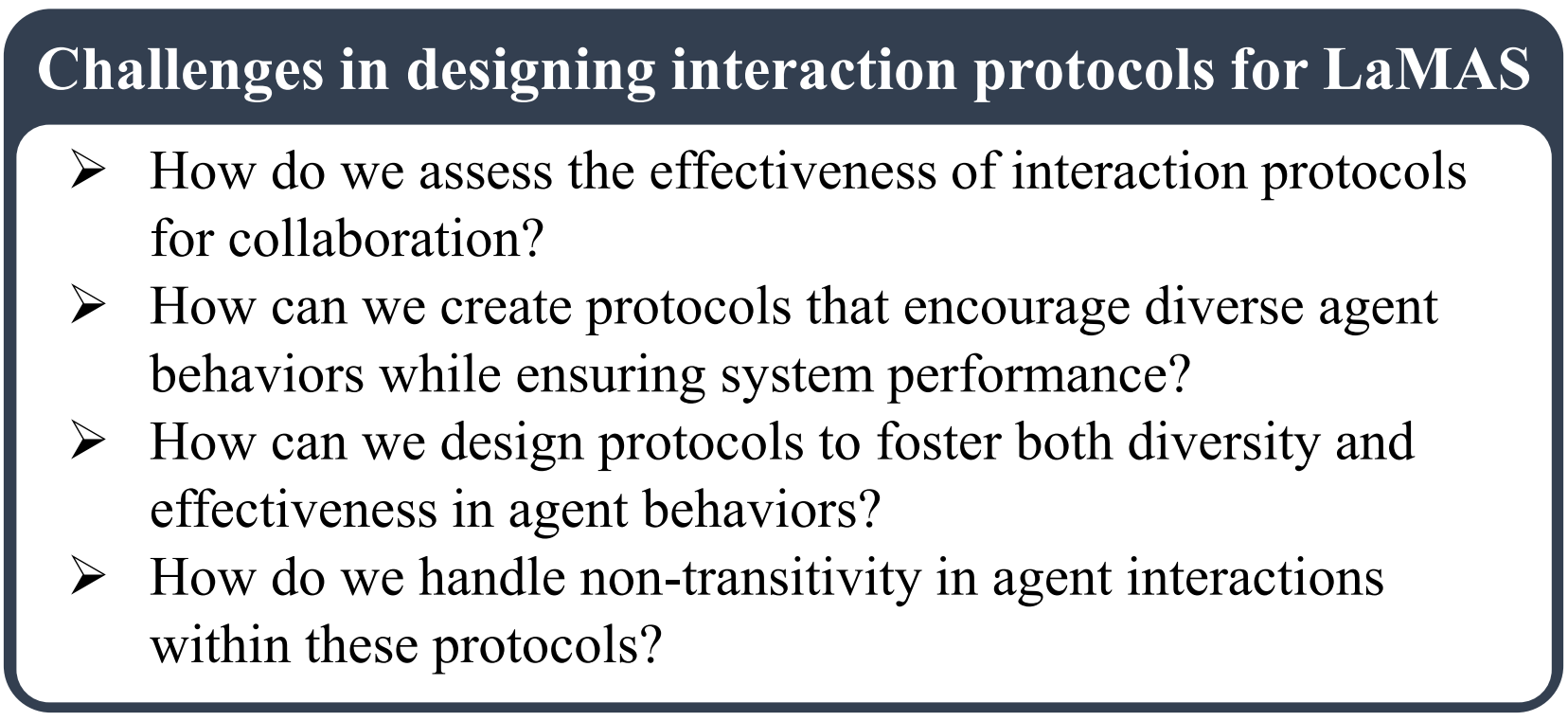}
    \label{fig:questions}
\end{figure}
\vspace{-0.5cm}

From these challenges, we identify 3 critical issues in protocol design. First, LaMAS requires a layered protocol architecture to manage diverse agent interactions efficiently, enabling dynamic protocol selection based on task and agent capabilities \cite{wooldridge2009introduction}. Second, as system scale increases, traditional protocols face limitations in managing communication overhead and maintaining consistency, necessitating innovative approaches to protocol design \cite{shoham2009multiagent}. Third, LaMAS leverages the strengths of LLM agents in language understanding and contextual interpretation. Protocols should leverage capabilities, such as handling ambiguous commands or enabling real-time negotiation to clarify information \cite{marro2024scalablecommunicationprotocolnetworks}. 

\minisection{Core Protocol Framework}
We propose a comprehensive protocol framework consisting of five essential components: the instruction processing protocol, the message exchange protocol, the consensus formation protocol, the credit allocation protocol and the experience management protocol.
\begin{figure}[t]
    \centering
    \includegraphics[width=0.92\linewidth]{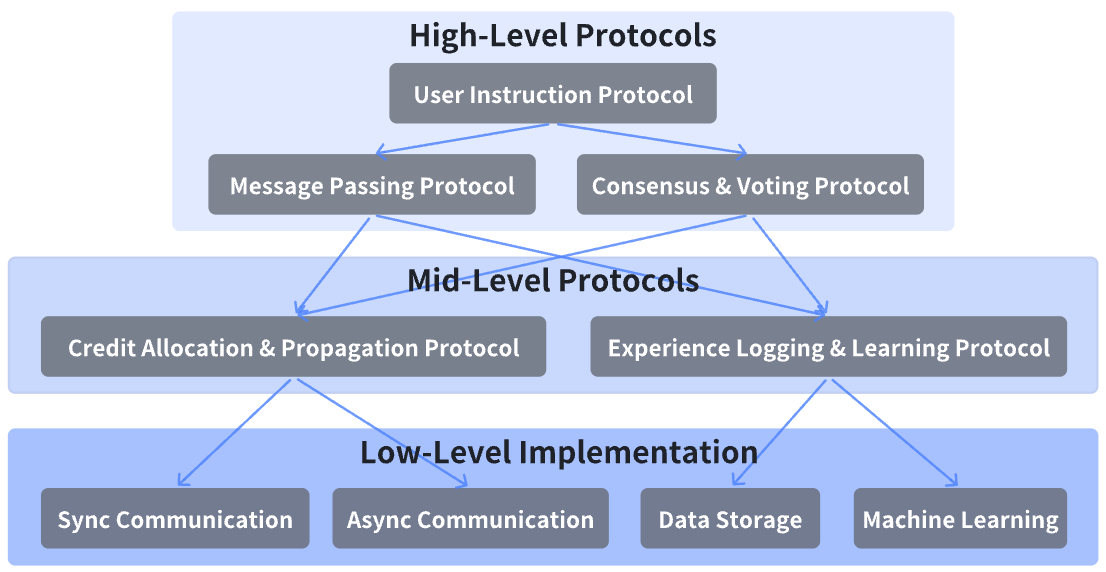}
    \vspace{-0.1cm}
    \caption{Protocol Hierarchy.}
    \label{fig:protocol_hierarchy}
    \vspace{-0.5cm}
\end{figure}

The Instruction Processing Protocol standardizes the interpretation of user instructions through structured parsing mechanisms and context-aware processing pipelines \cite{10.1145/1753171.1753181}. This protocol implements sophisticated disambiguation techniques to handle uncertain or incomplete instructions, maintaining consistency across multiple interaction rounds.

The Message Exchange Protocol establishes the foundation for inter-agent communication through standardized message formats and adaptive transmission mechanisms \cite{shoham2009multiagent, zhou2024symboliclearningenablesselfevolving}. This protocol dynamically switches between synchronous and asynchronous modes based on task requirements and system load, implementing priority-based routing algorithms to optimize message delivery under varying conditions.

The Consensus Formation Protocol implements distributed decision-making mechanisms through a combination of voting systems and negotiation frameworks \cite{10.1007/s10462-021-10097-x}. This protocol adapts consensus thresholds dynamically based on task criticality and system state, ensuring robust decision-making while maintaining system responsiveness. When proposals conflict, agents can resolve disagreements through negotiation, ensuring that progress is still made despite differences in opinion. Voting protocols allow agents to express preferences and reach a decision even when full consensus is not achievable, thus preventing deadlocks and ensuring that tasks continue to progress. 

The Credit Allocation Protocol addresses the challenge of fair contribution assessment through multi-level propagation mechanisms. Agents that participate in tasks receive corresponding credit based on their contributions \cite{zhou2024symboliclearningenablesselfevolving}. By implementing task-specific metrics and performance-based distribution algorithms, this protocol ensures equitable reward allocation while incentivizing collaborative behavior.

The Experience Management Protocol facilitates collective learning through structured logging and pattern extraction mechanisms \cite{zhang2024surveymemorymechanismlarge}. Each agent logs its experiences and learning outcomes during task execution. These experience records may include successes, failures, the effectiveness of strategies used, and interactions with other agents. This protocol implements cross-agent knowledge sharing algorithms, enabling systematic improvement of system performance through accumulated experience.

The effectiveness of LaMAS depends on the seamless integration of these protocols, as illustrated in Figure \ref{fig:protocol_hierarchy}. The hierarchical organization enables dynamic protocol selection and efficient resource utilization, while maintaining system scalability.

\subsection{Agent Training Methods}
In LaMAS, each agent has the incentive to improve itself in order to get more credits assigned to them from the platform. In terms of ``agent training", we mean the methods of improving the agent's performance, including tuning-free methods and parameter-tuning methods.

\minisection{Tuning-free Methods}
In the field of LLM agents, tuning-free methods are strategies to improve performance without modifying model parameters. These methods are beneficial when direct parameter tuning is costly or impractical. Key tuning-free methods include:
\vspace{-0.1cm}
\begin{itemize}[leftmargin=10pt]
    \item \textbf{Prompt Engineering}: Prompt engineering involves designing specific input prompts to elicit desired responses from LLM agents. By carefully structuring prompts, agents can align more closely with task requirements without any parameter adjustments \cite{brown2020language}.
    \item \textbf{Few-Shot Learning}: Few-shot learning provides limited examples within the prompt, helping the agent understand new tasks. In zero-shot learning, models tackle tasks solely through natural language instructions, leveraging pre-trained knowledge. Both approaches enable flexibility and adaptability in multi-agent environments \cite{radford2021learning, yao2023reactsynergizingreasoningacting, trad}.
    \item \textbf{External Tool Utilization}: Agents can enhance their capabilities by interacting with external tools or APIs. Using external resources, such as databases or calculators, allows agents to perform complex tasks without additional model training \cite{schick2023toolformerlanguagemodelsteach, patil2023gorillalargelanguagemodel}.
\end{itemize}

These tuning-free methods are particularly valuable in LaMAS. They enable agents to adapt quickly and work together on complex tasks in dynamic environments, supporting smooth collaboration with minimal computational costs.

\minisection{Parameter-tuning Methods}
To directly tune the parameters of the LLMs behind each agent, the alignment methods and multi-agent reinforcement learning methods can be used.
First, the alignment methods for tuning LLMs are generally based on supervised learning loss based on the target output or the preference of the experts. Directly fitting the experts' output corresponds to the behavioral cloning methods in agent imitation learning \cite{pomerleau1991efficient}, while training over the experts' preference pairs improves the agent's policy in a learning-to-rank manner \cite{rafailov2024direct}. Such a kind of method has not been much utilized in a multi-agent task as the alignment target is not clearly formulated in such a scenario.
Second, multi-agent reinforcement learning (MARL) is a key method for training agent policy in a multi-agent system, which formulates the task as a multi-agent sequential decision-making problem \cite{busoniu2008comprehensive}. Here, we mainly consider cooperative MARL methods as, in general, the agents are organized to pursue team success, i.e., to fulfill the user's task. According to the form of agent coordination in training and execution, MARL methods can be divided into three major categories, namely i) centralized training with centralized execution \cite{sukhbaatar2016learning,wen2022multi}, ii) centralized training with decentralized execution \cite{lowe2017multi,rashid2020monotonic}, and iii) decentralized training and execution \cite{tan1993multi,tian2019regularized}.

\subsection{Attacks and Defenses in LaMAS}
As LaMAS systems handle sensitive data and critical operations, their security is a top concern. The distributed nature of LaMAS introduces unique vulnerabilities beyond those of single-LLM systems. Malicious actors can target not only individual agents but also exploit inter-agent communications and collective decision-making processes. This section examines the attacks against LaMAS and the defense mechanisms, with a focus on their implications in multi-agent systems.

\minisection{Attack Surface and Vulnerabilities}
LaMAS face three main types of attacks. 

First, prompt injection attacks manipulate input prompts to trick models into generating harmful responses. These attacks are particularly dangerous in LaMAS, where compromised agents can propagate malicious prompts across the system. Recent work \cite{zou2023universal} show how slight changes in input phrasing can bypass defenses, while research \cite{liu2023prompt} demonstrate how system prompts can be modified using escape characters and context omission.

Second, memory and data poisoning attacks target the knowledge bases that agents use for decision-making. In LaMAS, poisoned data can affect multiple agents simultaneously. Research shows how contaminated knowledge bases in retrieval-augmented generation (RAG) systems can cause cascading errors throughout the agent network \cite{xue2024badrag}. Other study highlights how poisoned training samples with specific triggers can compromise fine-tuned agents, impacting system reliability \cite{yan2024backdooring}.

Third, model inversion and extraction attacks aim to reconstruct training data or extract model details through targeted queries. Analysis indicates that these attacks are particularly effective in LaMAS, where attackers responses from multi-agents to enhance extraction efficiency \cite{morris2023language}. The risk of data leakage is especially high for systems handling sensitive personal or commercial data, as shown by recent work on prompt-based vulnerabilities \cite{sha2024prompt}.

\minisection{Defense Mechanisms and Future Directions}
Several defense strategies have been proposed to counter these attacks, each addressing specific vulnerabilities in LaMAS environments.

Input sanitization techniques, such as prompt randomization and query encapsulation, help neutralize prompt injection attacks. One work demonstrate the effectiveness of these methods in LaMAS contexts, though they may introduce communication overhead \cite{robey2023smoothllm}. An improved approach involves adaptive delimiter strategies that maintain communication efficiency \cite{hines2024defending}.

Perplexity-based filtering is another promising defense. Several researches show that monitoring model perplexity can detect adversarial prompts without compromising model utility \cite{alon2023detecting, jain2023baseline}. In LaMAS, this method can be enhanced by cross-validating perplexity scores across agents, though careful calibration is required to avoid false positives during legitimate interactions.

Adversarially robust fine-tuning has also proven useful for enhancing LaMAS security. A dual-model approach, where adversarial samples are generated and validated during training, offers significant benefits \cite{o2023adversarial}. Further optimization focus on balancing robustness and utility, making these techniques particularly valuable for LaMAS by enabling system-wide application while preserving agent specialization \cite{xhonneux2024efficient}.

Despite these advances, challenges remain in securing LaMAS. Current defenses often struggle with the dynamic nature of agent interactions, where complex communication patterns can trigger false positives. Additionally, the computational overhead of comprehensive security measures can affect system performance, requiring a balance between security and efficiency.

Looking ahead, several research directions are crucial. First, there is a need for standardized security evaluation frameworks for LaMAS that account for individual agent vulnerabilities and system-wide risks. Second, developing lightweight security measures that maintain communication efficiency is an open challenge. Finally, adaptive defense mechanisms that evolve with emerging threats will be essential for long-term security.

To ensure the safe deployment of LaMAS, future research should focus on a holistic approach that combines robust model architectures, effective training procedures, and dynamic defense mechanisms. This will be critical for maintaining public trust as LaMAS systems continue to grow in complexity and impact.

\section{Key Business Aspects}
Drawing from our research on LaMAS, we present our vision of its business implications across three critical dimensions: \textbf{privacy preservation}, \textbf{traffic monetization}, and \textbf{intelligence monetization}. We anticipate how LaMAS could reshape business paradigms and explore the potential trajectories of its commercial applications.

\vspace{-0.2cm}
\subsection{Privacy Preservation in LaMAS}
The rise of LaMAS introduces new privacy challenges that go beyond those of traditional multi-agent systems. Unlike conventional agents, which exchange structured data, LLM agents handle rich, contextual information that may contain sensitive data embedded in natural language conversations, reasoning processes, and knowledge representations. Privacy preservation in LaMAS is critical because these systems process natural language data, which can inadvertently leak sensitive information through semantic connections and implicit knowledge representations.

\minisection{Privacy-Preserving Challenges}
We identify three levels of privacy concerns that require systematic analysis and novel solutions:
\vspace{-0.4cm}
\begin{itemize}[leftmargin=10pt]
    \item At the \textbf{semantic level}, the challenge lies in LLMs' natural language processing, which may inadvertently reveal sensitive information through contextual associations and semantic connections. Traditional privacy mechanisms, designed for structured data, are insufficient for handling such complex information, especially against attacks that exploit semantic vulnerabilities.
    \item At the \textbf{agent interaction level}, the continuous exchange of information between agents introduces privacy risks. Sensitive information can be exposed not only through direct content but also through behavioral patterns and response characteristics. Additionally, maintaining conversation history and context windows in each agent creates persistent vulnerabilities over time.
    \item At the \textbf{system architecture level}, the distributed nature of LaMAS complicates the enforcement of privacy guarantees across all components while maintaining system efficiency. The dynamic interactions between agents and their evolving knowledge further challenge the implementation of robust privacy protections.
\end{itemize}




\minisection{Privacy-Preserving Technologies}
Recent research has explored various technologies to address privacy challenges, ranging from cryptographic techniques to system-level solutions.

At the foundational level, Homomorphic Encryption (HE) \cite{cheon2017homomorphic} enables secure computation on encrypted data, supporting private agent-to-agent communication and inference. While promising in privacy-preserving machine learning \cite{gilad2016cryptonets} and secure data sharing \cite{chen2017fast}, HE’s computational complexity remains a significant challenge in LaMAS.

Secure Multi-Party Computation (SMPC) \cite{lindell2020secure} enables secure collaborative computation among multiple agents. SMPC has been used in privacy-preserving data analysis \cite{li2019privpy} and collaborative learning  \cite{knott2021crypten} in traditional multi-agent systems, but scaling this technology to large LaMAS remains known.

For system-level protection, Trusted Execution Environments (TEEs) \cite{costan2016intel} provide hardware-based security guarantees. Technologies such as Intel SGX \cite{costan2016intel}, ARM TrustZone \cite{pinto2019demystifying}, and AMD SEV \cite{sev2020strengthening} create secure enclaves for sensitive computations. However, integrating TEEs into LaMAS requires careful consideration of security and performance trade-offs.

Additionally, Differential Privacy (DP) \cite{dwork2006differential} offers mathematical methods for privacy-preserving data analysis. While effective for protecting sensitive information in collaborative tasks, DP faces unique challenges in natural language processing, such as managing privacy budgets and preserving utility.

\minisection{Research Directions and Open Challenges}
Advancing privacy preservation in LaMAS involves addressing several key issues. First, existing privacy metrics do not fully capture the complexities of semantic information leakage in natural language processing. There is a need for LaMAS-specific privacy frameworks that account for both direct and indirect information flows in semantic spaces. Second, integrating privacy-preserving technologies across LaMAS requires a unified approach that combines data protection, secure computation, and communication security. Performance optimization and scalability remain major hurdles, especially as agent networks expand, and maintaining privacy while enabling efficient collaboration is critical. Future research should focus on creating comprehensive privacy frameworks tailored to LaMAS. This includes standardizing privacy protocols, developing efficient implementations, and establishing evaluation metrics to ensure the practical deployment of privacy-preserving LaMAS systems.

\vspace{-0.2cm}
\subsection{Traffic Monetization}
Traffic Monetization in LaMAS involves generating commercial value by managing user traffic and optimizing ads, using the strengths of various agents. This includes improving traffic flow, boosting click-through rates (CTR), and increasing conversion rates (CVR) \cite{Kumar2016}. LaMAS leverages each agent’s capabilities to enhance user engagement and make advertising strategies more effective. The system also ensures fair and transparent revenue allocation. This section explores Traffic Monetization in LaMAS, covering business scenarios, revenue models, profit allocation, and the roles of application agents.

\begin{figure}[t]
    \centering
    \includegraphics[width=0.85\linewidth]{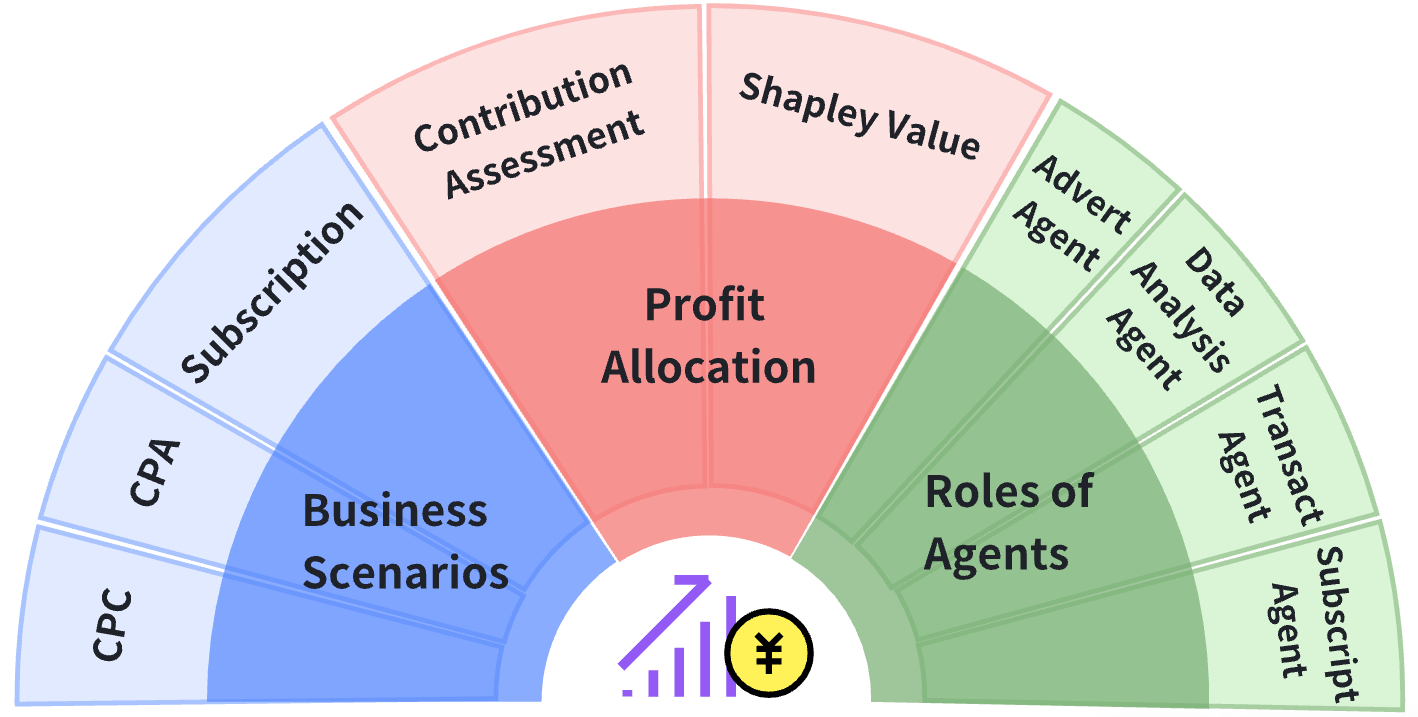}
    \caption{Traffic Monetization.}
    \label{fig:Traffic Monetization}
    \vspace{-0.3cm}
\end{figure}

\minisection{Business Scenarios and Revenue Generation}
In LaMAS, agents analyze user behaviors and preferences to optimize traffic management and deploy targeted advertisements. Through real-time data analytics, agents identify the most relevant ads to engage users, thereby enhancing click-through rates (CTR) and conversion rates (CVR). This approach builds user profiles and uses intelligent recommendation systems to personalize ads, boosting engagement and driving purchases. Revenue in LaMAS mainly comes from advertising, using Cost Per Click (CPC) and Cost Per Action (CPA) models \cite{Kannan2017}. In the CPC model, advertisers pay based on clicks, with agents earning commissions based on their contribution to traffic management and ad effectiveness. In the CPA model, payments are made for completed purchases, rewarding agents who drive conversions with a higher share of the revenue. In addition to advertising, income can be generated through user subscriptions for premium features or personalized services within specific applications, such as advanced analytics dashboards or exclusive access to specialized tools. These monetization strategies work together, with agents optimizing traffic flow, user engagement, and ad targeting to drive overall profitability.

\minisection{Profit Allocation Mechanisms}
Converting revenue into profits that are fairly distributed among agents is key to Traffic Monetization. The process starts by assessing each agent's contribution to traffic generation, ad clicks, and conversions, using metrics like CTR and CVR to quantify individual impact. To ensure fairness, LaMAS may use blockchain-based smart contracts to automate the distribution process, minimizing bias and human error. Additionally, a scoring system rates agents based on their performance, including factors like user feedback and engagement. Agents with higher scores receive a larger share of revenue, incentivizing better performance and continuous optimization. Attribution methods, such as the Shapley Value, ensure profits are allocated based on each agent’s contribution to the system \cite{shapley1953value}. Dynamic adjustment mechanisms allow real-time updates to revenue shares based on agent performance and market conditions, ensuring fair compensation for optimizing traffic and ads. Incorporating metrics like CPM (Cost Per Mille) adds a more nuanced approach to revenue allocation, offering a deeper understanding of ad performance beyond just clicks and conversions \cite{brynjolfsson2014second}.

\minisection{Roles of Application Agents}
Different types of application agents play distinct yet interrelated roles in Traffic Monetization. Advertising agents manage and deploy advertisements, using data analytics to optimize ad performance and increase user engagement. They select the best ad placements and adjust strategies based on real-time data and user behavior. Data analysis agents analyze user behavior, providing insights that help advertising agents refine ad strategies and improve content. These agents help create accurate user profiles and identify emerging trends to make advertising more effective. Transaction agents handle user purchases, ensuring smooth transactions and tracking conversions. They link sales performance to specific ads, helping advertisers improve future campaigns and boost conversion rates (CVR). Subscription agents manage premium services, offering personalized features and contributing additional revenue streams. These agents also contribute to long-term user engagement by enhancing retention and loyalty, supporting sustained revenue growth.

Traffic Monetization in LaMAS creates a system where revenue is generated and shared through the collaboration of different agents. By improving traffic management, optimizing ads, and ensuring fair distribution, LaMAS makes sure all agents benefit fairly. Future research should focus on improving attribution models like the Shapley Value to make profit allocation even fairer. Adding metrics like CPM will also help allocate revenue more accurately and improve monetization strategies.



\vspace{-0.3cm}
\subsection{Intelligence Monetization}
Intelligence Monetization in LLM-based Multi-Agent Systems represents a significant evolution in AI commercialization by leveraging the collaborative capabilities of specialized agents. Unlike traditional single-model paradigms, multi-agent systems enable dynamic interactions among specialized agents, each designed to address specific tasks, thereby facilitating the creation of more versatile and robust intelligence solutions. This paradigm has shown promise in various applications, as exemplified by Microsoft's Copilot Studio Platform, launched on November 19, 2024. The platform supports an ecosystem of over 1,800 large models and offers open APIs and integration tools, enabling enterprises to incorporate agent technology into workflows and applications for enhanced customization and scalability \cite{microsoft_ai_agent_ecosystem}. 

\minisection{Revenue Generation through Data-Driven Services}
A key revenue model in Intelligence Monetization within LaMAS is the sale of data-driven services \cite{SJODIN2021574, Li2018, HAJIPOUR2023100302, microsoft_ai_agent_ecosystem}. Specialized agents analyze distinct datasets—such as consumer preferences, product usage, and market trends—to generate actionable insights. These insights are delivered in the form of reports, forecasts, or tailored recommendations that businesses can purchase. For instance, one agent may provide personalized user behavior insights to enhance marketing, while another offers market trend analysis to guide strategic planning. By integrating specialized agents, LaMAS offers a broad range of insights, which can be monetized through subscription models or one-time reports. In practice, successful implementations like OpenAI's GPT-4 API \cite{openai2024gpt4technicalreport} have demonstrated how multiple specialized models can work in concert, with distinct agents handling different aspects of the intelligence pipeline. These include specialized agents for data processing and preprocessing, deep pattern recognition and insight generation, recommendation transformation, and platform integration.

\minisection{Innovative Licensing and Agent Marketplaces}
LaMAS also introduces novel licensing approaches that move beyond traditional software licensing. One of the most prominent is the Agent-as-a-Service (AaaS) \cite{Multi-Agent-as-a-Service}, as seen in Google Cloud’s AutoML, which enables dynamic agent deployment based on computational needs, with usage-based pricing and automatic scaling \cite{google_cloud_automl}. Complementing this will be the emergence of agent marketplace platforms, creating ecosystems for third-party agent development and deployment, as demonstrated by Hugging Face's model hub adapted for deployment of LLMs \cite{huggingface_model_hub}. Furthermore, hybrid deployment architectures will become increasingly popular, combining on-premise agent deployment for sensitive operations with cloud-based agents for scalable tasks, as seen in IBM's Watson services which use distributed agent architecture \cite{ibm_watsonx_assistant}.

Intelligence Monetization within LaMAS offers a sustainable revenue framework by harnessing the collective intelligence of specialized agents. Looking ahead, as demand for AI-driven insights grows, LaMAS’s role in delivering scalable, actionable intelligence will continue to expand. Future developments should focus on enhancing agent collaboration for real-time insights and adapting to emerging business models and industries, offering tailored solutions to meet specific needs.


\subsection{Integration of 3 Business Aspects}
The three key business aspects of LaMAS form an interconnected framework that drives both commercial success and ethical operation. Data privacy ensures trust and compliance while allowing secure data usage. This foundation supports traffic monetization by generating user engagement data while adhering to privacy regulations. Intelligence monetization transforms these privacy-preserved interactions into actionable insights and services. As LaMAS evolves, maintaining a balance between privacy, commercial success, and technological progress will be critical for long-term sustainability. Future development should focus on strengthening these connections while adapting to evolving privacy regulations and market demands.

\begin{figure}[t]
    \centering
    \includegraphics[width=0.8\linewidth]{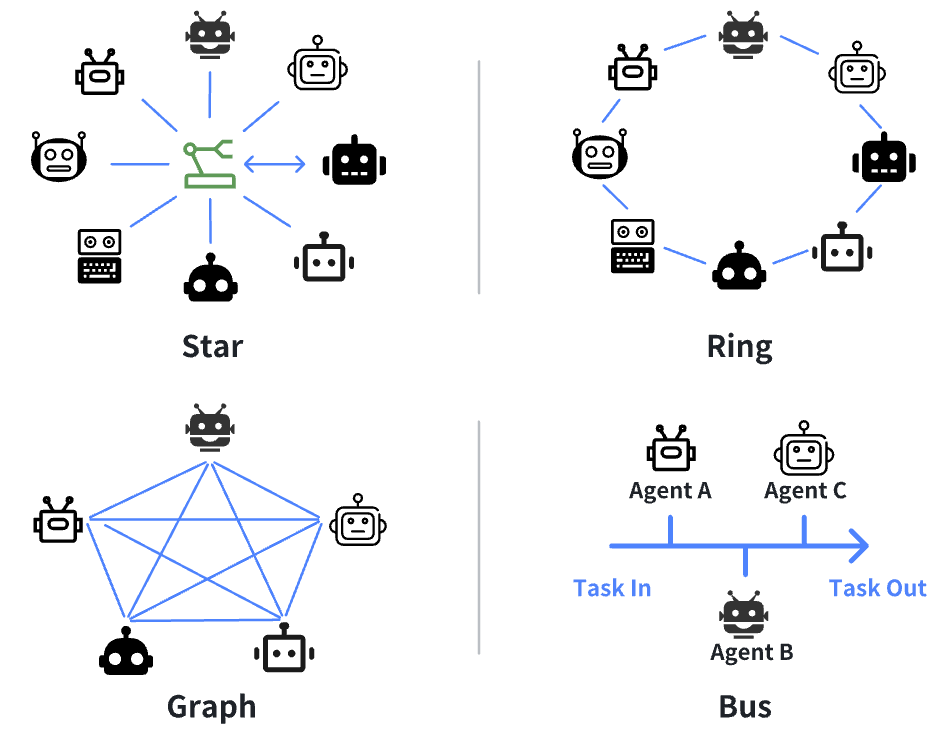}
    \vspace{-0.2cm}
    \caption{Architectures of LaMAS.}
    \label{fig:Architectures of LaMAS}
    \vspace{-0.3cm}
\end{figure}

\section{Case Study}
Building on the technical foundations and business considerations of LaMAS, we now delve into real-world implementations to illustrate how these theoretical frameworks are realized in practice. Through carefully selected case studies, we explore how different architectural choices influence system efficiency, data privacy, and monetization capabilities. These examples not only validate our theoretical insights but also shed light on the challenges and opportunities inherent in deploying LaMAS solutions.

\subsection{Architectures in LaMAS}
The implementation of LaMAS in real-world applications reveals various architectural patterns, each designed to address specific operational requirements and constraints. 

\begin{figure}[t]
    \centering
    \includegraphics[width=1\linewidth]{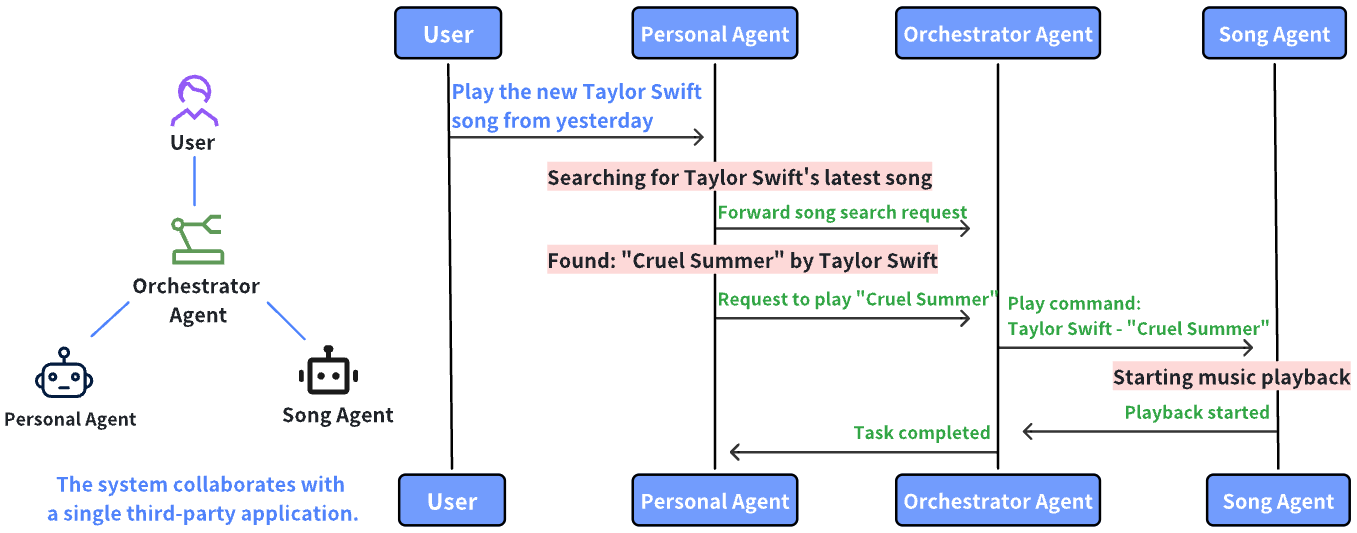}
    \caption{Centralized Architecture of LaMAS.}
    \label{fig:centralized architecture}
\end{figure}
\begin{figure}[t]
    \centering
    \includegraphics[width=1\linewidth]{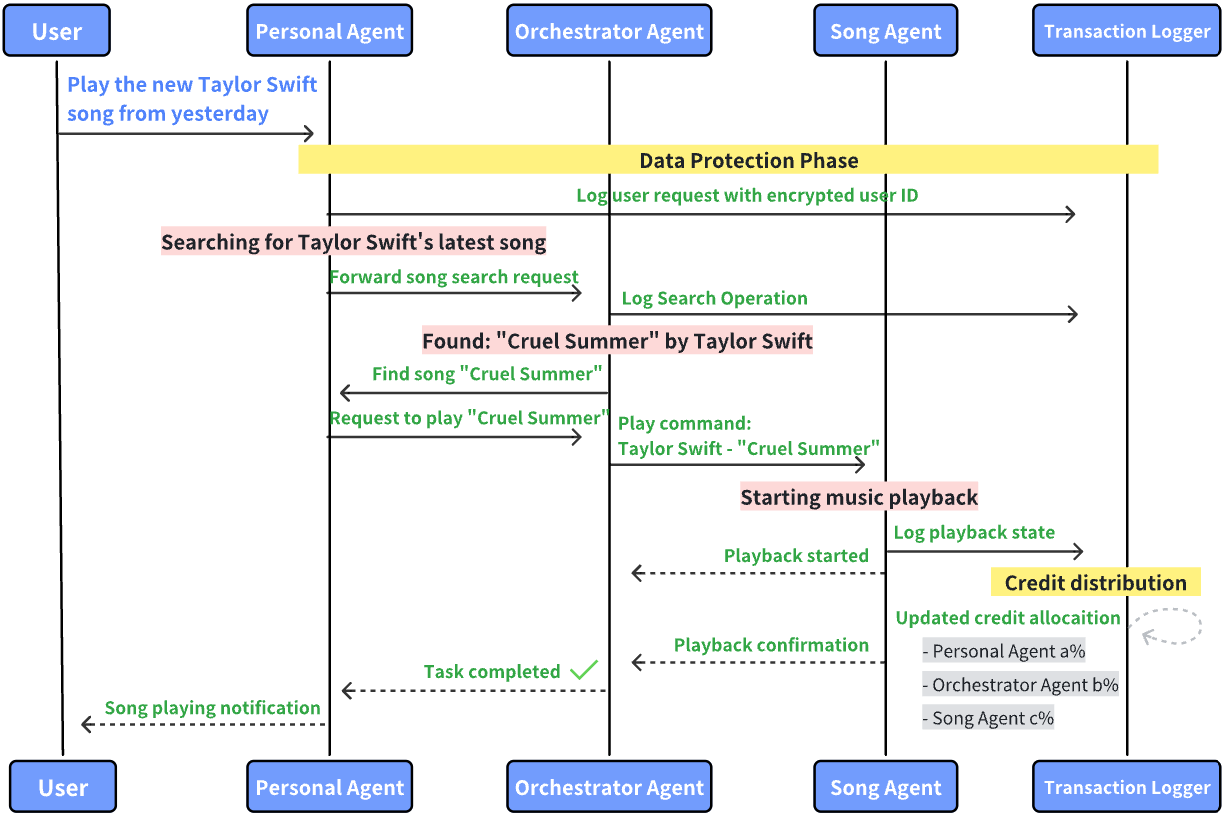}
    \caption{Centralized data handling of LaMAS.}
    \label{fig:Centralized data handling of LaMAS.}
\end{figure}

As we analyze current LaMAS deployments, we observe that architectural choices significantly influence system capabilities, from privacy protection to operational efficiency. Figure~\ref{fig:Architectures of LaMAS} illustrates four fundamental patterns that have emerged in practice:

\begin{itemize}[leftmargin=10pt]
    \item \textbf{Star Architecture}: In this structure, a central agent coordinates communication with all other agents \cite{chatdev,hong2024metagpt,wu2023autogen,xie2023openagents,fu2024msiagentincorporatingmultiscaleinsight}. This centralized control model works well when one agent is responsible for task distribution and overall orchestration.
    \item \textbf{Ring Architecture}: Agents are arranged in a circular configuration, each communicating with its predecessor and successor \cite{chan2023chatevalbetterllmbasedevaluators,liang2024encouragingdivergentthinkinglarge}. This decentralized structure supports sequential task processing, ensuring each agent has a specific role in the task pipeline.
    \item \textbf{Graph Architecture}: This network allows for a fully interconnected system where each agent can communicate directly with any other agent \cite{zhugegptswarm,chen2023agentverse}.This architecture creates a fully or non-fully interconnected network where each agent can communicate with their neighbors. It provides maximum flexibility and redundancy, allowing multiple communication pathways to support complex interactions.
    \item \textbf{Bus Architecture}: This structure uses a fixed workflow or Standard Operating Procedure (SOP), where tasks are sent to a central bus, which then distributes them to the appropriate agents or processes \cite{li2023metaagentssimulatinginteractionshuman,gao2024agentscopeflexiblerobustmultiagent,trirat2024automlagentmultiagentllmframework,li2024agentorientedplanningmultiagentsystems}. The bus ensures a clear input-output mechanism and a structured flow of tasks in a sequential manner. 
\end{itemize}

\subsection{A Decentralized Star Architecture in LaMAS}
Figure~\ref{fig:centralized architecture} and Figure~\ref{fig:Centralized data handling of LaMAS.} present our first case study of LaMAS implementation in a music service scenario. The system uses a centralized architecture where several agents, including a Personal Agent, an Orchestrator Agent, and a Song Agent, collaborate to process user's music playback requests. In this setup, the Orchestrator Agent acts as a central hub, managing communication and coordinating tasks among the agents.

However, this centralized approach means that all agents must send their data, including sensitive user information, through the Orchestrator Agent to complete tasks. While efficient for coordination, this design creates potential privacy and security risks since user data passes through multiple agents.

\begin{figure}[t]
    \centering
    \includegraphics[width=1\linewidth]{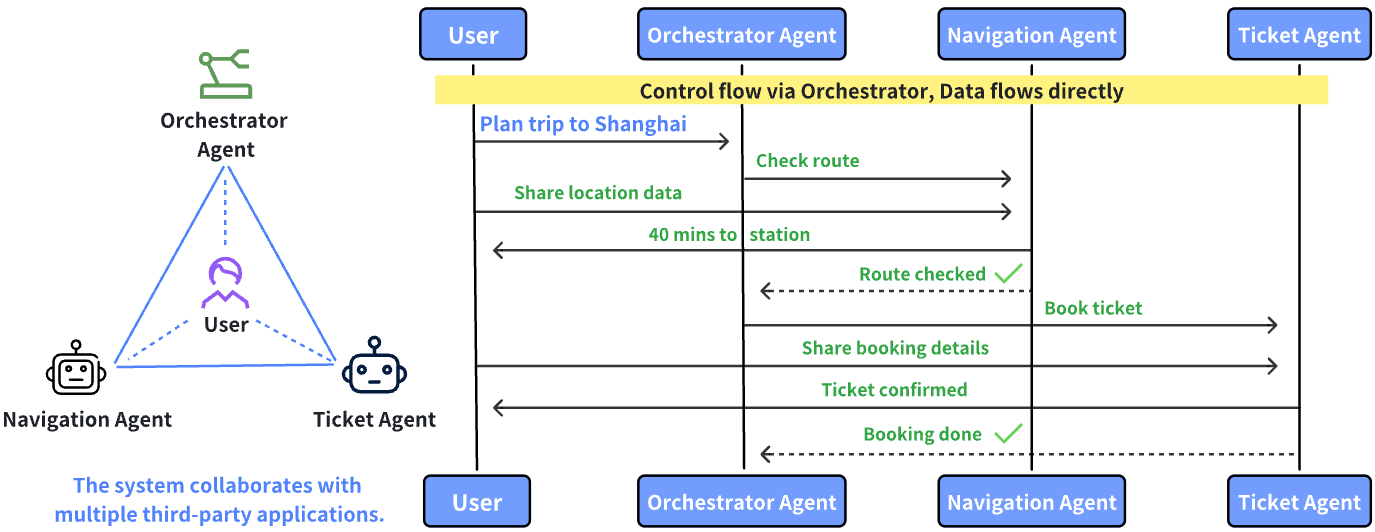}
    \caption{Decentralized Star Architecture of LaMAS.}
    \label{fig:decentralized architecture}
\end{figure}
\begin{figure}[t]
    \centering
    \includegraphics[width=1\linewidth]{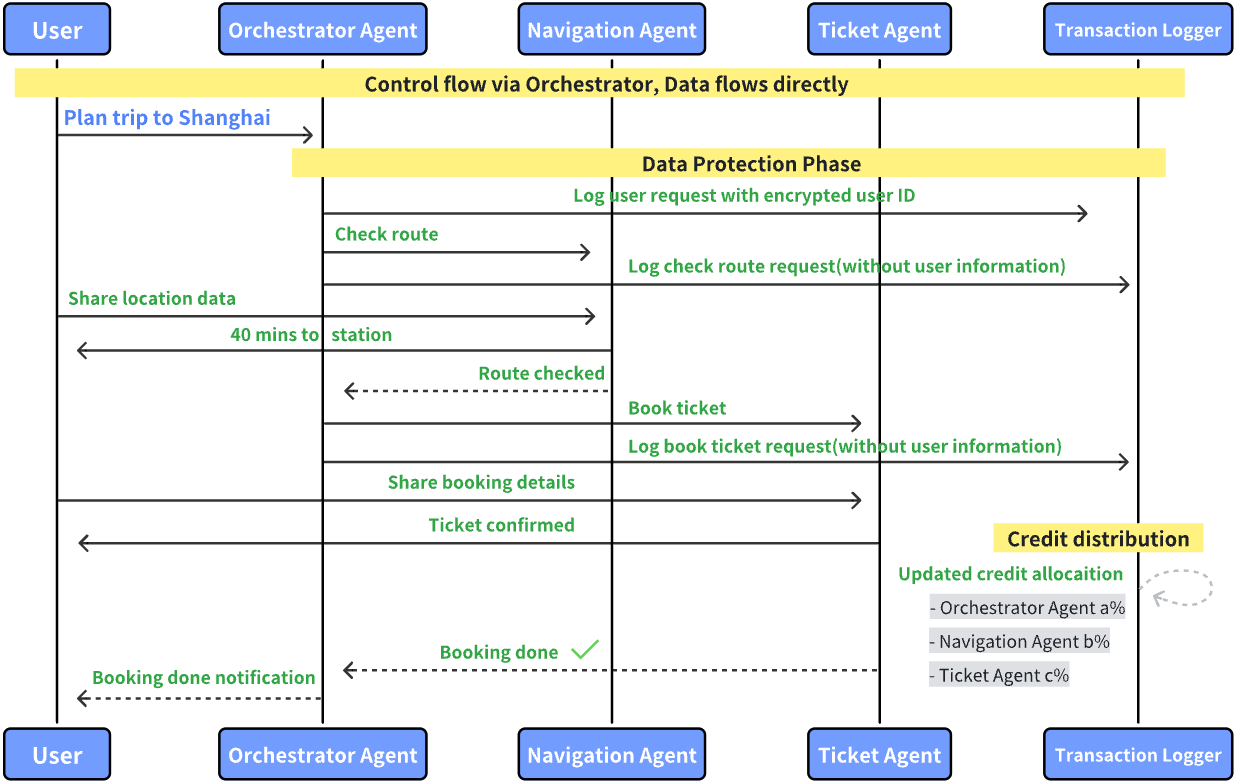}
    \caption{Decentralized data handling of LaMAS.}
    \label{fig:Decentralized data handling of LaMAS.}
\end{figure}

To address these privacy concerns, we propose a modified decentralized Star Architecture, illustrated in Figure~\ref{fig:decentralized architecture} and Figure~\ref{fig:Decentralized data handling of LaMAS.} through a travel booking scenario. In this new design, the Orchestrator Agent still coordinates tasks but avoids directly handling sensitive data. Instead, specialized agents, like the Navigation Agent or Ticket Agent, process their tasks independently and interact directly with user data when needed. This setup reduces privacy risks while maintaining system efficiency.

In the decentralized architecture, the Orchestrator Agent focuses on breaking down user instructions into smaller tasks and deciding the order of task execution. It stays uninvolved in sensitive data processing and only reconnects when tasks are completed or additional coordination is required. Each specialized agent handles its specific tasks within its own data domain, ensuring privacy and security.

A fair credit allocation system, as shown in Figure~\ref{fig:Centralized data handling of LaMAS.} and Figure~\ref{fig:Decentralized data handling of LaMAS.}, managed by the Transaction Logger, ensures that all agents receive appropriate rewards based on the tasks they complete and the resources they use. This credit allocation approach improves data protection and system security while keeping operations efficient.

\section{Conclusion \& Future}
In this paper, we provide our analysis on the future development of LLM-based Multi-Agent Systems (LaMAS) from the perspectives of techniques and business. Technically, compared to the traditional single-LLM-agent systems, LaMAS has a higher potential for overall performance and system flexibility; and commercially, LaMAS brings the feasibility of proprietary data preservability and monetization through traffic and intelligence, which essentially incentivizes various entities to contribute to the whole ecosystem. Several effective protocols for multi-agent communication and collaboration are developed in progress, which will drive the implementation of the LaMAS ecosystem towards achieving artificial collective intelligence in the near future.




\bibliographystyle{ACM-Reference-Format} 
\bibliography{main-folder/multi-llm-agents}


\end{document}